\documentclass[10pt,twocolumn,letterpaper]{article}

\usepackage{iccv}
\usepackage{times}
\usepackage{epsfig}
\usepackage{graphicx}
\usepackage{amsmath}
\usepackage{amssymb}

\usepackage[breaklinks=true,bookmarks=false]{hyperref}

\iccvfinalcopy 

\def\iccvPaperID{8508} 

\ificcvfinal\fi

\begin{document}
\renewcommand{\thefootnote}{\fnsymbol{footnote}}

\title{Cascade Image Matting with Deformable Graph Refinement}
\author{Zijian Yu \thanks{Joint first authors.}\\
School of Software, BNRist, \\Tsinghua University\\
Beijing, China\\
{\tt\small zj-yu19@mails.tsinghua.edu.cn}
\and
\setcounter{footnote}{0}
Xuhui Li \footnote{}\\
School of Software, BNRist, \\Tsinghua University\\
Beijing, China\\
{\tt\small  lixh20@mails.tsinghua.edu.cn}
\and
Huijuan Huang\\
Kuaishou Technology\\
Beijing, China\\
{\tt\small  huanghuijuan@kuaishou.com}
\and
Wen Zheng\\
Kuaishou Technology\\
Beijing, China\\
{\tt\small  zhengwen@kuaishou.com}
\and
Li Chen\\
School of Software, BNRist, \\Tsinghua University\\
Beijing, China\\
{\tt\small  chenlee@tsinghua.edu.cn}
}

\maketitle
\ificcvfinal\thispagestyle{empty}\fi

\begin{abstract}
   Image matting refers to the estimation of the opacity of foreground objects. It requires correct contours and fine details of foreground objects for the matting results. To better accomplish human image matting tasks, we propose the Cascade Image Matting Network with Deformable Graph Refinement, which can automatically predict precise alpha mattes from single human images without any additional inputs. We adopt a network cascade architecture to perform matting from low-to-high resolution, which corresponds to coarse-to-fine optimization. We also introduce the Deformable Graph Refinement (DGR) module based on graph neural networks (GNNs) to overcome the limitations of convolutional neural networks (CNNs). The DGR module can effectively capture long-range relations and obtain more global and local information to help produce finer alpha mattes. We also reduce the computation complexity of the DGR module by dynamically predicting the neighbors and apply DGR module to higher--resolution features. Experimental results demonstrate the ability of our CasDGR to achieve state-of-the-art performance on synthetic datasets and produce good results on real human images.
\end{abstract}

\section{Introduction}
Image matting refers to the problem of extracting high--quality alpha mattes (the opacity of foreground object at each pixel) from a set of given images. 
As a practical image processing technology, matting has a variety of applications for image and video editing.
Generally, the composition of image $\mathbf{I}$ is expressed as a linear equation as follows:
\begin{equation}
\label{eq1}
\mathbf{I}_i=\alpha_i\mathbf{F}_i+(1-\alpha_i)\mathbf{B}_i,  \alpha_i\in[0,1],
\end{equation}
where $\mathbf{I}_{i}$ is the RGB color at pixel \emph{i}, $\alpha_{i}$ is the matte value at pixel \emph{i} and  $\mathbf{F}_{i}$ and $\mathbf{B}_{i}$ are the RGB colors of the foreground and background at pixel \emph{i}. Matting is a highly ill-posed problem, i.e., there are seven unknown values and only three known values at each pixel, which increases the difficulty of solving matting problems. Although existing works have provided effective ways to perform matting, they still present limitations.

The first limitation is that most existing works ~\cite{DIM2017, AdaM2019, CAM2019, FBAM2020, LFM2019} have predicted alpha mattes by using a one-pass encoder--decoder network, which may result in inaccurate contours and artifacts when foreground and background have similar local features. This is mainly due to that those methods predict alpha mattes from single--scale features and cannot make full use of the global and local information contained in the image.


The second limitation is that existing CNN-based matting methods cannot well handle certain slender objects (e.g., human hair). In addition to the basic use of CNN in the network architecture, some matting works have tried to refine the details of alpha mattes after the backbone network with CNN-based module. Xu \etal~\cite{DIM2017} used a lightweight fully convolutional neural network (CNN) to generate sharp boundaries for alpha mattes. Cai \etal~\cite{AdaM2019} proposed a propagation unit that could refine alpha mattes with accurate details and less artifacts.
However, these CNN-based refinement methods are restricted by the fixed shape of convolutional kernels and a limited receptive field, leading to performance degradation when manipulating slender objects.

To overcome the first limitation, we simulate the matting logic of human. While meeting the matting tasks, people generally first determine the overall contour of the foreground object and then iteratively improve the details in boundary areas under the guidance of the contour. 
Therefore, we design a network cascade architecture for image matting to generate more accurate contours and details of foreground objects. Our method predicts coarse alpha mattes from low--resolution images as contours and then progressively supplements the details from high--resolution images under the guidance of the contours. Through this low-to-high, coarse-to-fine pipeline, our network can supplement local information with global information and estimate extremely finer alpha mattes with correct contours and precise details. 


To overcome the limitations of CNN and produce better performance on slender objects, we apply a graph neural network(GNN) to extract features with higher quality.
Compared with CNN, GNN has shown its ability to better capture long-range dependencies from data.
Some existing works~\cite{3DGNN2017, SuperPoints2019, DGMPN2020} used GNN to improve the performance of detection and segmentation. 
However, these GNN-based methods are limited by high computation complexity and time consumption, as GNN requires large number of nodes and thus can only be applied to low--resolution feature maps or superpoints set obtained by clustering.   

Inspired by deformable convolutional networks~\cite{DeformConv2017} that can dynamically adjust kernel shapes according to objects, we propose the Deformable Graph Refinement (DGR) module to reduce the computation cost of graph construction and propagation. The DGR module uses the convolutional network to predict the coordinates of neighbors and performs information aggregation and transmission among pixels. 

We combine two solutions above and propose a method called the Cascade Image Matting Network with Deformable Graph Refinement (CasDGR). First, the network cascade architecture is designed to enhance the  simulation of the coarse-to-fine matting logic. Second, the DGR module is adopted to improve the obtaining of more appropriate features and the handling of slender objects.

The main contributions of this study are as follows:
\begin{itemize}
    \item We propose an end-to-end automatic image matting approach to produce high-quality alpha mattes from single RGB images.
    \item We design a network cascade architecture to estimate alpha mattes in a coarse-to-fine manner.
    \item We present a Deformable Graph Refinement module based on GNN that can preserve more details of the matting results and be applied on higher--resolution features.
\end{itemize}


Some existing work~\cite{DIM2017, IM2019, GCAM2020} require trimaps as additional inputs. However, the construction of high--quality trimaps is complicated. Automatic matting methods~\cite{LFM2019, BGM2020, HAttM2020, BSHM2020} whose inputs do not contain trimaps are more challenging, but more convenient and feasible for some applications, such as matting for human only. Our CasDGR is also automatic and can achieve good matting performance with single RGB images. 
Similar to~\cite{BGM2020}, we test our method on Adobe human image dataset~\cite{DIM2017}. 
The experimental results demonstrate that our method can achieve  state-of-the-art performance and produce excellent visual results. What's more, our automatic matting approach outperforms existing trimap-based methods both quantitatively and qualitatively. We also test the CasDGR on natural human images. Our method shows good performance on real-world human images as well. 


\section{Related Work}

\subsection{Image Matting}

Current image matting methods can be divided into traditional methods and learning-based methods.

\textbf{Traditional methods.} 
Sampling-based methods~\cite{BM2001,SharedSamplingM2010,ComSamplingM2013,GlobalSM2011,OCSRM2007} mainly use statistical methods to sample and model the color of known foreground and background regions and determine the best color pair of each unknown pixel and calculate the alpha mattes. Propagation-based methods~\cite{KNNM2013,CFM2004,PM2004,NLM2011,SpectralM2008} 
propagate the alpha values of known regions to unknown regions according to the affinities among adjacent pixels. 
Nevertheless, traditional methods utilize color information and location information instead of semantic information and context information, which may lead to loss of essential detail.

\textbf{Learning-based methods.} 
Learning-based matting methods compensate the disadvantages of traditional methods and generally offer better performance. 
Trimap-based learning methods require annotated trimaps as additional inputs. 
Cho \etal~\cite{DCNN2016} utilized the results of~\cite{KNNM2013} and ~\cite{CFM2004} and normalized RGB color to predict alpha mattes by using a deep CNN.
Xu \etal~\cite{DIM2017} first proposed an encoder--decoder structure network to estimate alpha mattes. The refinement stage in their work could produce extrmely sharp boundaries. 
Hou \etal~\cite{CAM2019} used two encoders to extract local and global context information and perform matting.
Cai \etal~\cite{AdaM2019} adopted a multi-task learning method to complete two subtasks, and a propagation unit was used to process the results of the two subtasks and consequently obtain the final alpha mattes.
Forte and Piti{\'{e}}~\cite{FBAM2020} proposed to predict foregrounds, backgrounds, alpha mattes by using a single encoder--decoder. 
Hao \etal~\cite{IM2019} optimized the upsampling operator and applied it to image matting.
Tang \etal~\cite{LSM2019} utilized sampling networks and a matting network to perform color sampling and matting. 

Automatic methods do not need additional trimaps, which hence the avoidance of constraints of trimaps.
Shen \etal~\cite{DAPM2016} estimated trimaps by using a CNN and performed matting with method of ~\cite{CFM2004}.
Sengupta \etal~\cite{BGM2020} used
disturbed backgrounds and segmentation results as additional inputs to simultaneously predict alpha mattes and foregrounds. 
Zhang \etal~\cite{LFM2019} first obtained the probability maps of a foreground and a background and
then fused them to obtain the final alpha mattes. 
Liu \etal~\cite{BSHM2020} used coarse annotated data coupled with fine annotated data to improve matting performance.
Qiao \etal~\cite{HAttM2020} used channel and spatial attention mechanisms to extract multi-level features from a set of single images. 

Most deep learning methods aim to enhance the matting based on the single encoder--decoder architecture 
and do not provide effective refinement refinement stage.
We apply the network cascade architecture to our CasDGR to perform the matting process in a coarse-to-fine manner. A novel DGR module is proposed for feature refinement. The experimental results prove that both proposed techniques can achieve a considerably improvement in the matting results.

\subsection{Network Cascade}

Network cascade is an effective architecture for many computer vision tasks, such as detection~\cite{FPN2017,CasRCNN2018}, segmentation~\cite{DASS2017}, and pose estimation~\cite{CPN2018}. The central idea of using the network cascade is to solve challenging tasks in a 
coarse-to-fine manner. 
Cai \etal~\cite{CasRCNN2018} presented Cascade R-CNN to achieve progressive refinement of detection results. 
Chen \etal~\cite{CPN2018} predicted multiple heatmaps of human keypoints in high-to-low resolution and fused them using RefineNet. 
Li \etal~\cite{DASS2017} handled easy regions in shallow layers and hard regions in deep layers to improved the accuracy and speed of semantic segmentation.
To the best of our knowledge, the CasDGR is an early attempt to adopt a network cascade architecture into image matting tasks.

\subsection{Graph Neural Network}

Many graph neural networks (GNN)~\cite{GNN2009, GGNN2017, DCNN2016, GCN2017, GraphSAGE2017, GAT2018} have been proposed to solve the general problems of graphs.
Compared with CNN, GNN can better capture long-range dependencies from data, which benefits many computer vision tasks, such as detection~\cite{PointGNN2020,CasGNN2020,RepGNN2020}, segmentation~\cite{3DGNN2017,SuperPoints2019,DGMPN2020,RepGNN2020} , and pose estimation~\cite{PGNN2019,3dpose2020}.
Luo \etal~\cite{CasGNN2020} designed Cascade-GNN for RGBD salient object detection to exploit useful information from RGB and depth images.
Cai \etal~\cite{3dpose2020} used the graph convolution network to exploit the spatial and temporal relationship of 3D human body and hand pose. 
In ~\cite{3DGNN2017, SuperPoints2019, PointGNN2020}, the authors proposed GNN-based methods for segmentation and detection on 3D point clouds. However, the methods manifest restrictions on data size resulting from the high computation cost and low running speed of GNN.
DGMN~\cite{DGMPN2020} and RepGNN~\cite{RepGNN2020} can reduce the computation cost by dynamically sampling the nodes, consequently improving the performance of segmentation and detection.
Our work can predict the neighbors of each node and adopted into higher resolution feature maps, thus helping to obtain more details to solve the image matting problems.

\begin{figure*}
\begin{center}
    \includegraphics[width=0.9\linewidth]{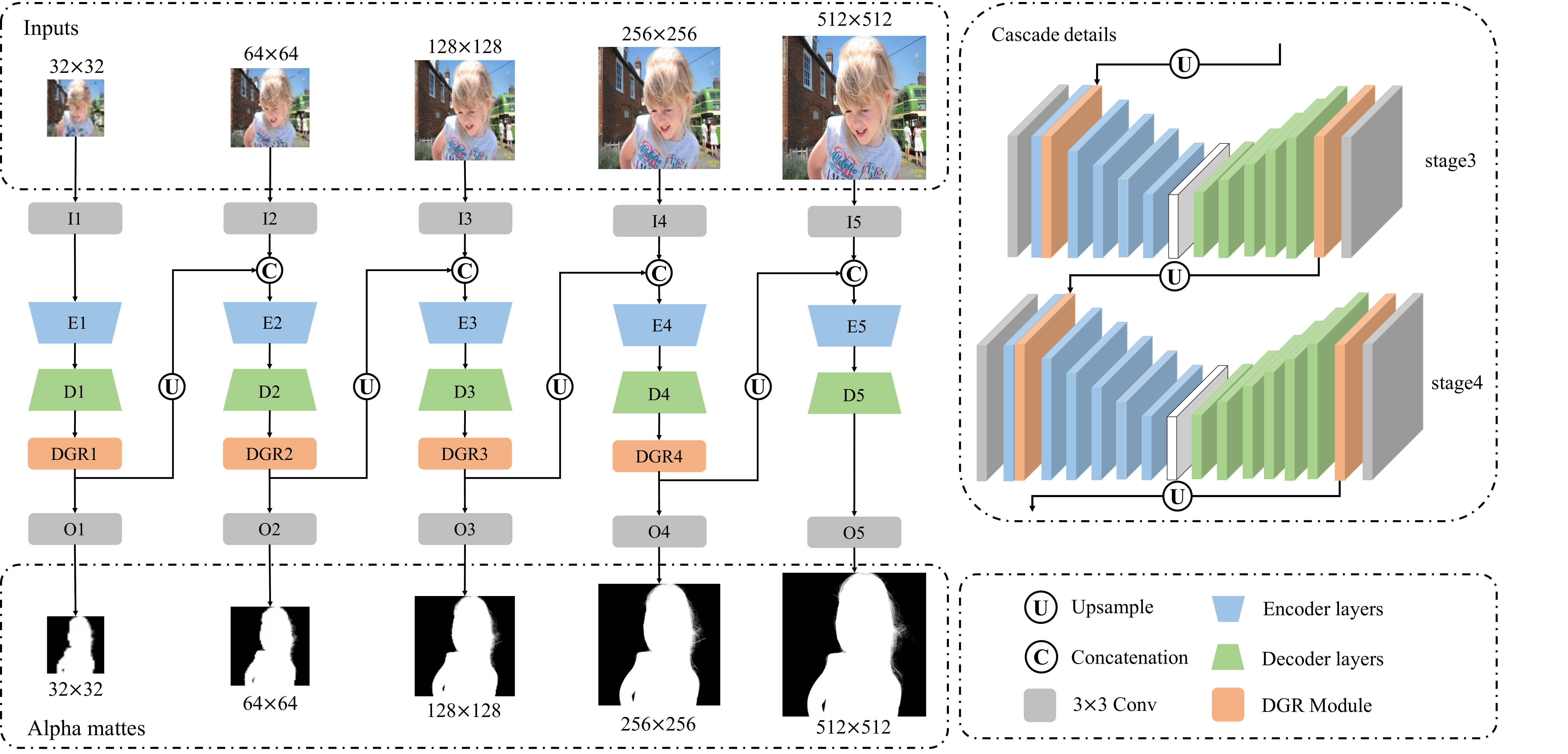}
\end{center}
   \caption{Overview of the proposed CasDGR. The main architecture is a cascade network contains 5 stages, where each stage is an encoder--decoder network followed by a DGR module. Given an input image, we downsample it to multi-scale inputs for each stage and estimate multi--resolution alpha mattes from low to high. We only use the predicted alpha matte of the last stage for further evaluation. }
\label{fig:overview}
\end{figure*}
\section{Approach}
In this section, we first introduce the overall network architecture and details of our CasDGR. Then, the loss functions and implementation details are presented.

\subsection{Cascade Network Design}

As shown in Figure~\ref{fig:overview}, the central idea of our approach is to use a network cascade architecture to predict alpha mattes from low to high resolution. The CasDGR consists of five stages in total. Similarly to most previous works, each stage is composed of an encoder--decoder U-structure network. Inspired by $\rm U^2$-Net~\cite{U2Net2020}, we use Residual U-block (RSU) as the backbone network in each stage owing to its ability to extracting multi-scale features and its low computation cost. 
The input of each stage contains an image with different resolutions scaled from the original image.
No other additional inputs are required by our network. In stage $m$ ($m\in[2,4]$), we first use a 3$\times$3 convolutional layer to generate the 64-channel feature map $\mathbf{F}_{in}^m$ from the input image. Then the $\mathbf{F}_{in}^m$ is concatenated with the refined feature map 
$\mathbf{F}_{re}^{m-1}$ from the previous stage. The RSU block takes the concatenation of two feature maps as the input and then outputs feature map $\mathbf{F}_{out}^m$ with the same resolution as $\mathbf{F}_{in}^m$. $\mathbf{F}_{out}^m$ is fed into the Deformable Graph Refinement (DGR) module to generate the 64-channel refined feature map $\mathbf{F}_{re}^m$.
Finally, a 3$\times$3 convolutional layer is used to predict the 1-channel alpha mattes from $\mathbf{F}_{re}^m$. Moreover, $\mathbf{F}_{re}^m$ is upsampled two times for concatenation with $\mathbf{F}_{in}^{m+1}$ in the next stage. 

The details of the RSU blocks are also shown in Figure~\ref{fig:overview}. The encoder part continuously performs convolution and downsampling on the feature map, whereas the decoder part performs upsampling and convolution to restore the feature map to the original resolution. 
Skip connections are applied to corresponding layers between the encoder and decoder. 
Atrous convolution is used in the deep layers to further enlarge the receptive field. Different from original RSU blocks, we use group normalization (GN)~\cite{GN2018} instead of batch normalization (BN) after each convolutional layer in our network, 
because our CasDGR is trained with small a batch size (2 on each GPU). Furthermore, the performance of BN may degrade when the batch size is small.

The CasDGR handles image matting tasks in a coarse-to-fine manner and predicts multiple alpha mattes from low to high resolution. In the earlier stages, the network extracts more global information in concordance with the much larger receptive field from the downsampled input image. This approach can help to improve the detection of the foreground object area. The predicted alpha mattes from these earlier stages can be regarded as coarse segmentation masks of the foreground object in visual perception. In the later stages, the predicted alpha mattes are further improved by using both higher--resolution input images and feature maps from previous stage; the former supplements the detail that may have been lost in earlier stages, whereas the latter provides rich semantic information. The DGR module further improves the quality of the generated feature maps by means of a graph-based model, which will be discussed in the succeeding sections.
Thus, our CasDGR can progressively refine the details from stage 1 to 5 while maintaining the correct contour of the foreground correct and consequently produce high--quality alpha mattes. 

\subsection{Deformable Graph Refinement}
\begin{figure}
\begin{center}
    \includegraphics[width=1\linewidth]{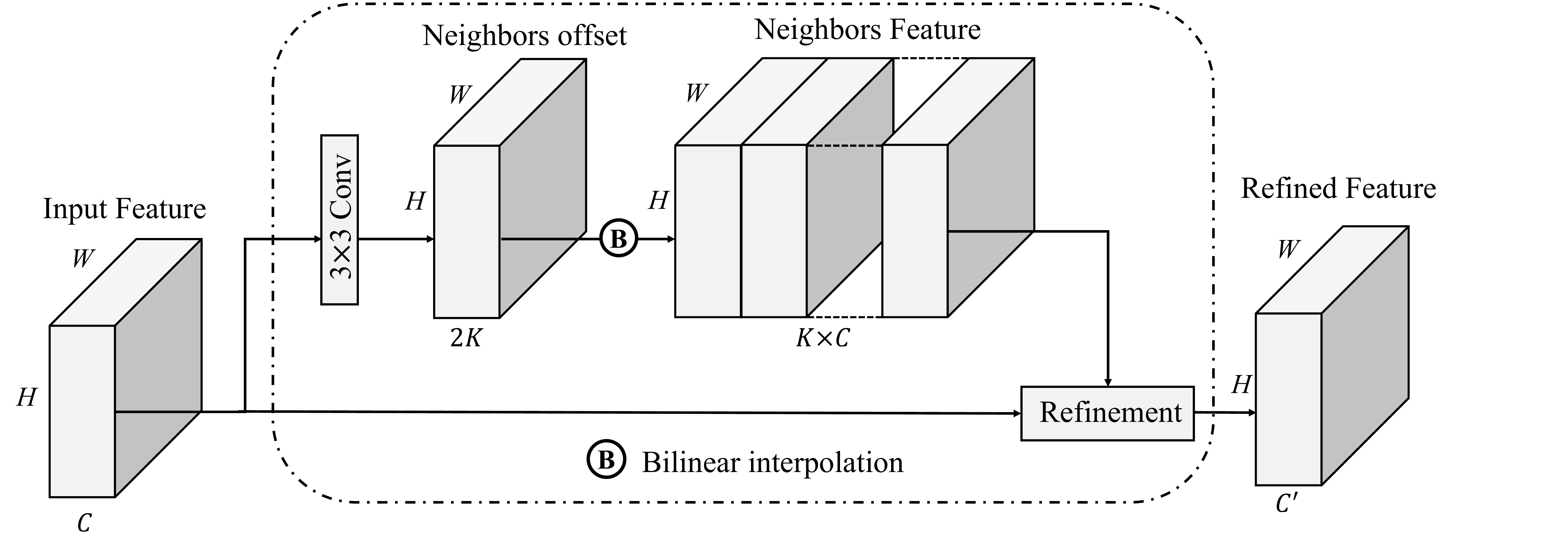}
\end{center}
   \caption{Illustration of the Deformable Graph Refinement (DGR) module. The input is the feature map output by the RSU block. DGR first predicts $K$ neighbor coordinates for each node and calculates feature values of neighbors. Then DGR updates feature values of each node by a refinement stage.}
\label{fig:GRM}
\end{figure}

We propose the Deformable Graph Refinement (DGR) module for feature map refinement. The details of the DGR module are shown in Figure~\ref{fig:GRM}. We regard the feature map with a shape of $H\times W\times C$ as a composition of $H\times W$ nodes and construct a graph on them in which each node entails a $C$-dimension feature.
The DGR module is inspired by deformable convolutional networks~\cite{DeformConv2017}, which dynamically adjust convolution kernels. 
We assume that each pixel in the feature map $\mathbf{F}_{out}\in\mathbb{R}^{H\times W\times C}$ outputed from thedecoder initially has $K$ adjacent neighbors initially and use a convolutional layer to apply a 2D offset to each neighbor. Then, we calculate the neighbor coordinates and use the bilinear interpolation method to obtain neighbor the feature values from $\mathbf{F}_{out}$. We design a model for the neighbors' information aggregation and the feature map refinement. For a node $i$ in $\mathbf{F}_{out}$, we refine its feature as follows:
\begin{equation}
    \label{gnn:calw}
    s_{ij}=(\mathbf{W}_1\mathbf{F}_{out}^i)^T(\mathbf{W}_2\mathbf{F}_{out}^j),j\in\mathcal{N}(i),
\end{equation}

\begin{equation}
    \label{gnn:softmax}
    \beta_{ij} = \frac{\exp{(s_{ij})}}{\sum_{k\in\mathcal{N}(i)}\exp{(s_{ik})}},
\end{equation}

\begin{equation}
    \label{gnn:sum}
    \mathbf{F}_{re}^i = \sigma(\sum_{j\in\mathcal{N}(i)}\beta_{ij}\mathbf{W}_2\mathbf{F}_{out}^j),
\end{equation}
, where $\mathcal{N}(i)$ is the neighbors set of node $i$. $\mathbf{W}_1,\mathbf{W}_2\in\mathbb{R}^{C^{'}\times C}$ are two weight matrices that can be optimized. Eq.  \ref{gnn:calw} calculates the similarity $s_{ij}$ between node \emph{i} and its neighbor \emph{j}. 
Then, Eq. \ref{gnn:softmax} is calculated with a softmax function to normalize $s_{ij}$. The $\beta_{ij}$ after normalization are the weights of neighbor \emph{j} for node \emph{i}. Finally, in Eq. \ref{gnn:sum}, we aggregate the features of all neighbors with different weights. $\sigma$ is the ReLU activation function.
This feature refinement stage can be performed iteratively. By using the DGR module, our network can capture long-range relations between the distant pixels. DGR can also reduce the computation complexity and time consumption of graph construction by dynamically predicting the neighbors. We apply the DGR module to stages 1 to 4 in the cascade network and the highest resolution of the feature map refined by DGR can reach 256$\times$256, which is higher than those in the previous works~\cite{3DGNN2017, RepGNN2020}. We use the refined feature maps for feature connection and alpha prediction.
\subsection{Loss Functions}
In the training process, we use supervision at each stage of the CasDGR. Our loss function is defined as follows:
\begin{equation}
\label{L_o}
\mathcal{L}=\sum_{m=1}^{M-1}\lambda_a^m\mathcal{L}_a^m+\mathcal{L}_a^M+\lambda_c\mathcal{L}_c^M+\lambda_g\mathcal{L}_g^M,
\end{equation}
, where $\mathcal{L}_a^m$($M$ = 5 represents five stages) is the alpha prediction loss between the output alpha of stage $m$ and the labels with the same resolution, $\mathcal{L}_c$ is the compositional loss, and $\mathcal{L}_g$ is the gradient loss. We use all three losses for the last stage and only use alpha prediction loss for the previous stages. $\lambda_a^m$, $\lambda_c$, and $\lambda_g$ are the weights of each loss item. We use the normalized L1 loss to calculate all three losses:
\begin{equation}
\label{L_a}
\mathcal{L}_a^m=\frac{1}{|\Omega|}\sum_{i\in\Omega}||\hat{\alpha}_i^m-\alpha_i^m||_1,
\end{equation}
, where $\alpha_i^m$ is the predicted alpha values of stage $m$ at pixel $i$, $\hat{\alpha}_i^m$ is the ground truth alpha values resized to the same resolution as $\alpha^m$ at pixel $i$, and $|\Omega|$ is the number of pixels in $\alpha_i^m$ and $\hat{\alpha}_i^m$.
\begin{equation}
\label{L_c}
\mathcal{L}_c=\frac{1}{|\Omega|}\sum_{i\in\Omega}||\mathbf{I}_i-\alpha_i \mathbf{F}_i-(1-\alpha_i)\mathbf{B}_i||_1,
\end{equation}
, where $\mathbf{I}$ is the input image combined by foreground $\mathbf{F}$, background $\mathbf{B}$, and ground truth alpha matte, similiar to those in Eq. \ref{eq1}. $\alpha$ is the predicted result of the last stage.
\begin{equation}
\label{L_g}
\mathcal{L}_g=\frac{1}{|\Omega|}\sum_{i\in\Omega}||\nabla \hat{\alpha}_i-\nabla \alpha_i||_1,
\end{equation}
, where $\nabla \hat{\alpha}$ and $\nabla \alpha$ represent the normalized gradient of the predicted alpha and the ground truth alpha.

The training process aims to minimize the $\mathcal{L}$ of Eq. \ref{L_o}. $\mathcal{L}_a, \mathcal{L}_c$ can improve the pixel-wise accuracy of the predicted alpha mattes, and $\mathcal{L}_g$ is beneficial the production of highly  precise boundaries. We choose the predicted results of the last stage as the final of the output alpha mattes.

\subsection{Implementation Details}

We implement CasDGR by using PyTorch. In the training process, all images are randomly cropped to a resolution between 512$\times$512 and 800$\times$800 and then resized to 512$\times$512. For the data augmentation, we adopt horizontally random flipping together with brightness, contrast, and saturation augmentation on each training pair to avoid overfitting. We downsample the 512$\times$512 images to lower resolutions and feed them into the different stages of our method. The training set is shuffled at each epoch. For the group normalization layer in our network, the input feature map is separated into several 32-channel groups. 
We train our network from scratch until the loss converges. All convolutional layers in RSU blocks are initialized using the  Xavier method~\cite{Xavier2010}. Parameters of the  $3\times3$ convolutional layers in the DGR module are initialized to zero. The adam optimizer is used for loss optimization, with the initial learning rate set to 1e-4 and the other hyper parameters set to default. We clip the predicted alpha values of each stage to 0 to 1 for loss calculation and set $\lambda_a^m=\lambda_c=\lambda_g=1$ in Eq. \ref{L_o} in the experiments.

During testing, the input images are resized to 512$\times$512 before feeding them into the network. We evaluate different metrics between the 512$\times$512 predicted alpha mattes and ground truth. We train our CasDGR on 2 RTX 3090 GPUs with a batch size of 4. Only about 1 day are needed for the network to converge on the training set.

\section{Experiments}

In this section, we compare our approach with existing matting methods on the Adobe human image dataset, which is collected from the Adobe Composite-1k Dataset~\cite{DIM2017}. We show the quantitative and visual results of all testing methods and perform ablation studies on our CasDGR to demonstrate the importance of essential architectures and components in our method.

\subsection{Dataset and Evaluation Metrics}

\textbf{Dataset.} 
Adobe Composite-1k Dataset~\cite{DIM2017} contains 431 foreground images for training and 50 foreground images for testing with high--quality alpha annotations. Following Sengupta's work~\cite{BGM2020}, we use a subset of 280 images in the experiments (269 images for training and 11 images for testing). We filter the semi-transparent objects in the dataset to closely simulate the data distribution to the human matting scene in the real world. For the training set, each foreground image is combined with 100 background images from the COCO dataset~\cite{COCO2014}. For the testing set, each foreground image is combined with 20 background images from the PASCAL VOC2012 dataset~\cite{PASCAL2010}.

\textbf{Evaluation metrics.} 
We use four common metrics in image matting to evaluate the predicted alpha mattes, namely sum of absolute differences (SAD), mean squared error (MSE), gradient error (Grad), and connectivity error (Conn). Generally, the SAD and MSE metrics are more focused on numerical differences, whereas the Grad and Conn metrics proposed by~\cite{Metrics2009} pay more attention to the visual
perception of human observers.

\subsection{Ablation Study on the Adobe Testing Dataset}

To verify the role of some architectures and components of our method, we completed the ablation studies discussed below by using the Adobe testing dataset.

\textbf{Ablation on DGR.} Table~\ref{ab_dgr} shows the influence of different number of neighbors and iteration times of refinement stage on matting performance. Compared with Cascade network without DGR, CasDGR with different settings can improve all four metrics on on Adobe testing dataset. We find that only considering 1 neighbor can increase four evaluation metrics effectively. As the number of neighbors increases form 1 to 5, test results are improved too. However, further increasing the number of neighbors to 9 will lead a decline of matting performance.

\begin{table}
\begin{center}
\begin{tabular}{|l|cccc|}
\hline
Model & SAD & MSE & Grad & Conn \\
\hline
Ours-Baseline & 3.78 & 0.0065 & 4.67 & 3.73 \\
Ours-Cascade & 2.92 & 0.0046 & 2.85 & 2.77 \\
\hline
CasDGR$_{K=1,1-layer}$ & 2.25 & 0.0025 & 2.45 & 2.10 \\
CasDGR$_{K=1,2-layer}$ & 2.05 & 0.0021 & 2.16 & 1.88 \\
CasDGR$_{K=5,1-layer}$ & 1.93 & 0.0018 & 1.95 & 1.74 \\
CasDGR$_{K=5,2-layer}$ & \textbf{1.76} & \textbf{0.0015} & \textbf{1.66} & \textbf{1.54} \\
CasDGR$_{K=9,1-layer}$ & 2.16 & 0.0023 & 2.30 & 1.99 \\
CasDGR$_{K=9,2-layer}$ & 1.84 & 0.0017 & 1.79 & 1.63 \\
\hline
\end{tabular}
\end{center}
\caption{Ablation study on DGR module. Ours-Baseline: 1-stage network. Ours-Cascade: 5-stage cascade network without DGR. CasDGR: cascade network with DGR, $K$ means the number of neighbors in DGR, and ${layer}$ means the number of iterations.}
\label{ab_dgr} 
\end{table}

\begin{figure}
\begin{center}
    \includegraphics[width=1\linewidth]{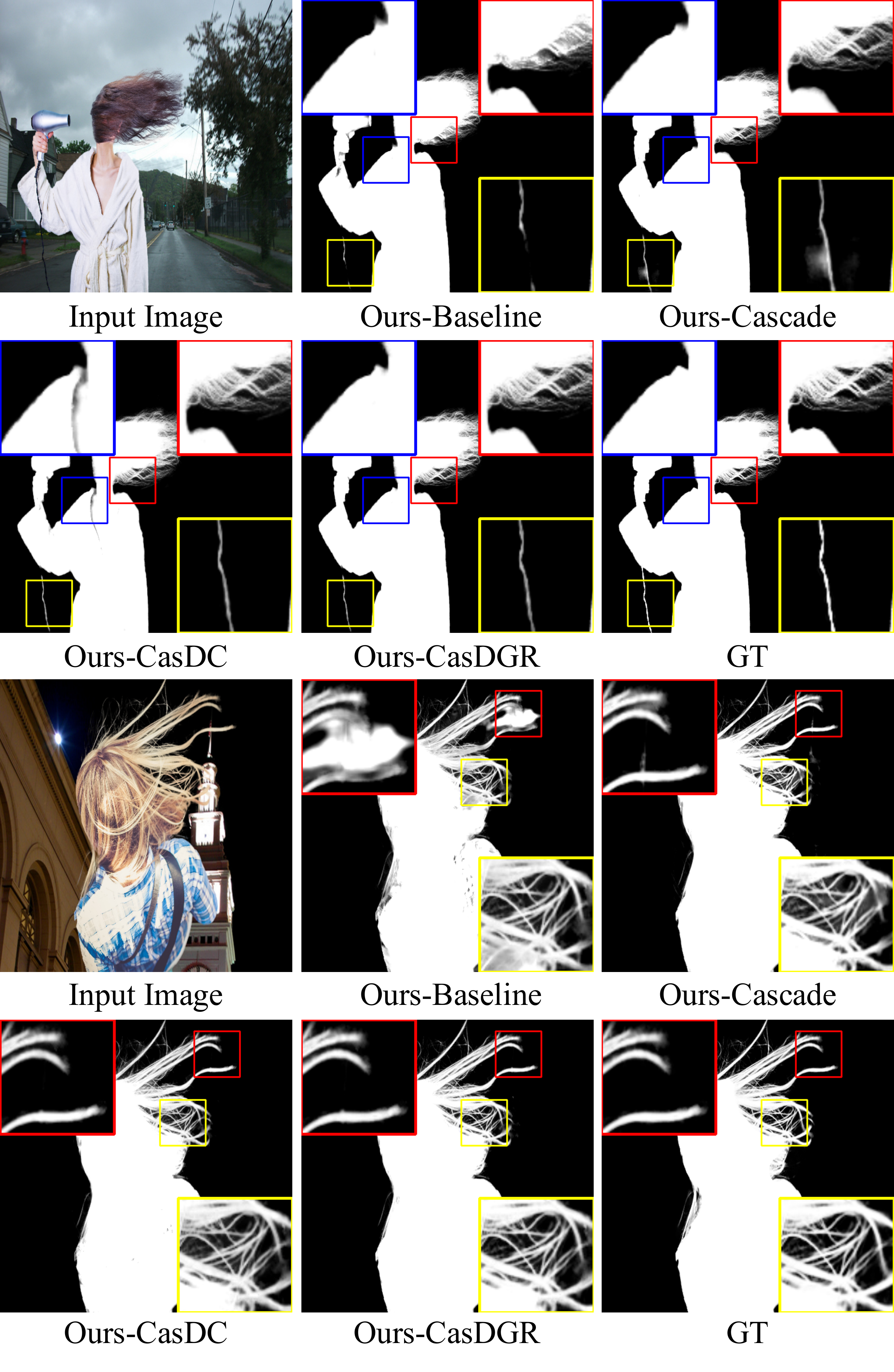}
\end{center}
   \caption{Visual results of ablation studies.}
\label{fig:ablation}
\end{figure}

\begin{table}
\begin{center}
\begin{tabular}{|l|c|c|c|}
\hline
Model & Ours-CasDCN & \multicolumn{2}{|c|}{Ours-CasDGR} \\
\hline
$layer$ & 1 & 1 & 2 \\
\hline
SAD & 2.13 & 1.93 & 1.76 \\
MSE & 0.0023 & 0.0018 & 0.0015 \\
Grad & 2.27 & 1.95 & 1.66 \\
Conn & 1.93 & 1.74 & 1.54 \\
\hline
Flops(G) & +8.71 & +5.65 & +8.36 \\
Params(M) & +0.57 & +0.12 & +0.16 \\
Inference Time(ms) & 51.23 & 41.33 & 48.56 \\
\hline
\end{tabular}
\end{center}
\caption{Ours-CasDCN vs Ours-CasDGR($K=5$). The value of FLOPs and Params are the increased value in contrast to Ours-Cascade. The results are measured with $512\times512$ input size on one GeForce RTX 3090 card. The batch size is 1.}
\label{table2}
\end{table}

Increasing the iteration times of refinement stage is also beneficial to image matting. For different settings of $K$, CasDGR with 2 iterations produce better results than 1 iteration. More iterations will increase the time and memory consumption as well. To balance the efficiency and effectiveness of the model, we choose $K=5$ and set iteration times to 2 as default in other experiments.

\textbf{Role of Network Cascade Architecture.} As shown in Table~\ref{ab_dgr}, the cascade network has achieved substantially improvement on all metrics compared with the the baseline network, which only uses the 1-stage network for matting. According to the visual results in Figure~\ref{fig:ablation}, Ours-Baseline produces some artifacts in the results, whereas Ours-Cascade can generate more visually accurate alpha mattes. The network cascade architecture has effectively improved the quantitative and visual results for matting.

\textbf{Role of the DGR.} Table~\ref{ab_dgr} has shown the improvement of DGR on evaluation metrics. In terms of the visual results in Figure~\ref{fig:ablation}, Ours-CasDGR can further refine the results compared with Ours-Cascade, which reduces some artifacts and is completed with more detail for the alpha mattes.

In addition, certain details in Figure~\ref{fig:ablation},  can be used to clearly analyze the matting refinement process of our method. In the case of the image of women with hand-held hair dryers, Ours-Baseline does not predict the complete wire of hair dryer in the lower left corner. After the network cascade, Ours-Cascade can predict a relatively complete wire, but some artifacts around it are apparent. Finally, after the refinement processing of Ours-CasDGR, the artifacts are removed, and a complete and fine wire is obtained. This step-by-step refinement process also verifies the design ideas and feasibility of our method.

\begin{figure*}
\begin{center}
    \includegraphics[width=0.95\linewidth]{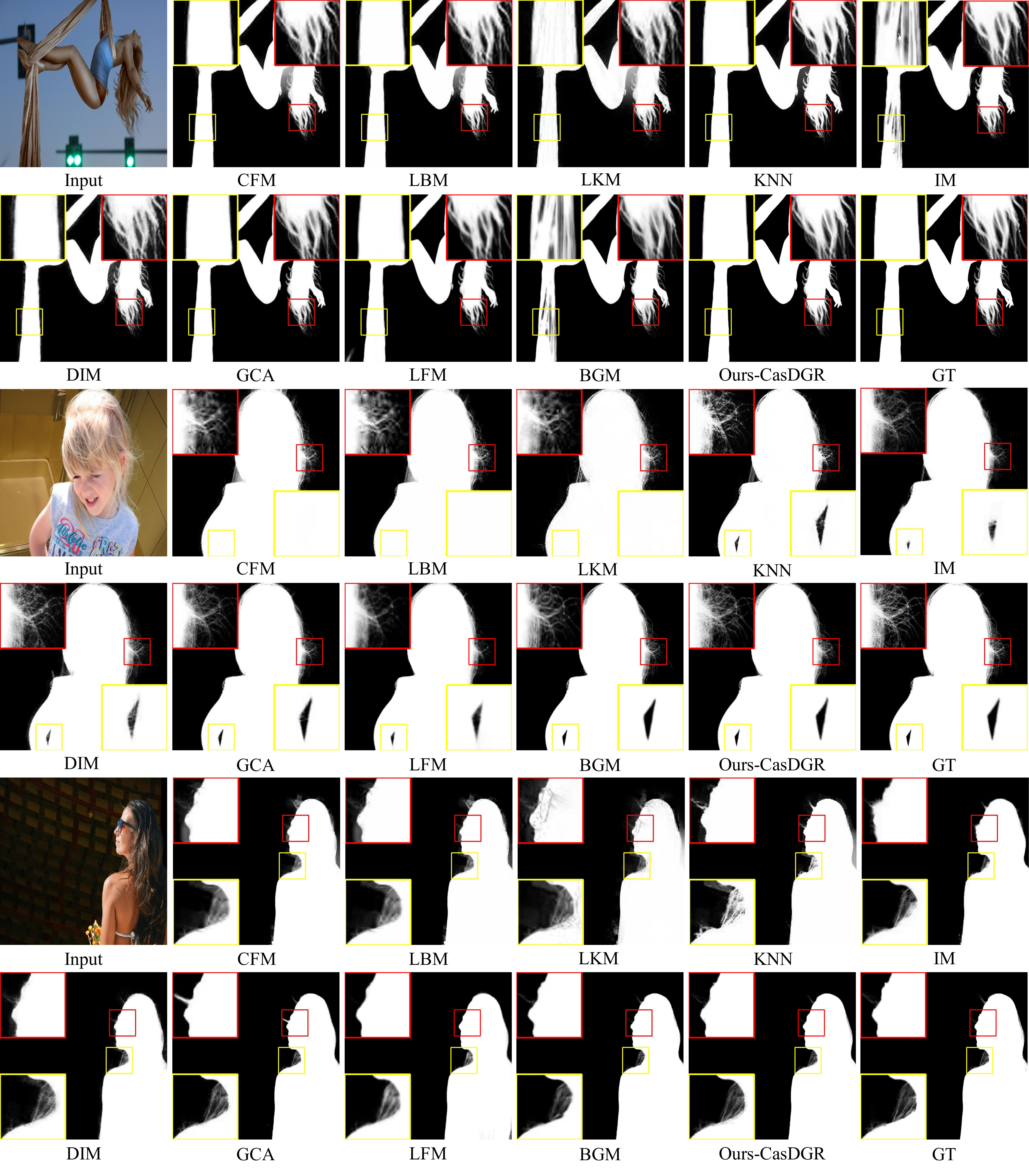}
\end{center}
   \caption{The visual results on Adobe testing dataset.}
\label{fig:result_cmp}
\end{figure*}

\textbf{Comparision with the DCN.} We demonstrate the superiority of DGR module by replacing the DGR in our matting network with deformable convolutional networks~\cite{DeformConv2017}. As shown in Table~\ref{table2}, Ours-CasDGR outperforms Ours-CasDCN on all metrics. According to the comparison of some model attributes, Ours-CasDGR can achieve better matting results with less Flops, Params, and inference time compared with Ours-CasDCN, which further demonstrates the improvements from using our approach. The visual results in Figure~\ref{fig:ablation} show that Ours-CasDGR is also visually superior to Ours-CasDCN, based on the finer alpha mattes of the former method.

\subsection{Comparison on the Adobe Testing Dataset}

We compare our approach on the constructed Adobe human image dataset with different kinds of available approaches.  \textbf{The traditional methods}: Closed-Form Matting (CFM)~\cite{CFM2004}, Learning Based Matting (LBM)~\cite{LBM2009}, KNN Matting (KNNM)~\cite{KNNM2013}, Random Walks Matting (RWM)~\cite{RWM2005}, and Large Kernels Matting (LKM)~\cite{LKM2010}. \textbf{The trimap-based learning methods}: Deep Image Matting (DIM)~\cite{DIM2017}, IndexNet Matting (IM)~\cite{IM2019}, and Guided Contextual Attention Matting (GCAM)~\cite{GCAM2020}. \textbf{The automatic learning methods}: Late Fusion Matting (LFM)~\cite{LFM2019} and Background Matting (BGM)~\cite{BGM2020}. 

\begin{table}
\begin{center}
\begin{tabular}{|l|cccc|}
\hline
Method & SAD & MSE & Grad & Conn \\
\hline
CFM~\cite{CFM2004} & 3.48 & 0.0040 & 3.87 & 3.35 \\
LBM~\cite{LBM2009}  & 3.68 & 0.0047 & 4.17 & 3.65 \\
KNNM~\cite{KNNM2013} & 3.73 & 0.0044 & 3.90 & 3.67 \\
RWM~\cite{RWM2005} & 4.96 & 0.0093 & 10.56 & 4.93 \\
LKM~\cite{LKM2010} & 5.52 & 0.0053 & 5.32 & 4.65 \\
\hline
\hline
IM~\cite{IM2019} & 2.29 & 0.0022 & 2.51 & 2.06 \\
DIM~\cite{DIM2017} & 2.58 & 0.0025 & 2.93 & 2.42 \\
GCAM~\cite{GCAM2020} & 1.89 & 0.0017 & 1.99 & 1.68 \\
\hline
\hline
BGM~\cite{BGM2020} -  \emph{Seg, $B^{\prime}$} & 2.30 & 0.0025 & 2.34 & 2.10  \\
BGM~\cite{BGM2020} -  \emph{Seg, $B$} & 2.28 & 0.0024 & 2.29 & 2.08  \\
LFM~\cite{LFM2019} & 4.35 & 0.0067 & 4.01 & 3.98 \\
Ours-CasDGR  & \textbf{1.76} & \textbf{0.0015} & \textbf{1.66} & \textbf{1.54} \\
\hline
\end{tabular}
\end{center}
\caption{Results on the Adobe testing dataset. 
\emph{Seg, $B^{\prime}$, $B$}: coarse segmentation results, disturbed backgrounds with Gaussian noises, and original backgrounds for Background-Matting~\cite{BGM2020}. 
}
\label{table1} 
\end{table}

During the evaluation, we resize input images to 512$\times$512 resolution to inference the alpha mattes and compute four metrics between the predicted alpha mattes and ground truths. For the approaches requiring trimaps, we resize the original trimaps in Adobe dataset to 512$\times$512 resolution as additional inputs. 
As BGM~\cite{BGM2020} needs segmentation results and disturbed backgrounds as additional inputs, we generate the segmentation results by applying person segmentation~\cite{personseg2018} and adding erosion (5 iterations), dilation (10 iterations) and a Gaussian blur ($\sigma$ = 5). We also generate the disturbed backgrounds by adding Gaussian noises $\eta\sim\mathcal{N}$ ($\mu$ = 3, $\sigma$ = 3) to the original backgrounds. The manner
of generating segmentation results and disturbed backgrounds are the same as those in BGM~\cite{BGM2020}. 

The quantitative results are shown in Table~\ref{table1}. The implications of our experimental results are as follows:

Our CasDGR can achieve state-of-the-art results on all metrics among all testing approaches on Adobe testing dataset, i.e., the traditional methods, trimap-based, and automatic methods mentioned above. The experimental results demonstrate that our approach can achieve the best human matting performance by using a single input image.



Our CasDGR outperforms other matting methods especially on the Grad and Conn metrics. As Grad and Conn focus more on the visual effects of human observers, the comparison results indicate that the CasDGR can achieve great matting performance in visual perception, which is also proven by the visual results in Figure~\ref{fig:result_cmp}.



As shown in Figure~\ref{fig:result_cmp}, our CasDGR has a high--quality visual effect on human images and can preserve fine contour and detail of the foreground object. Although GCA~\cite{GCAM2020} and BGM~\cite{BGM2020} can also generate precise alpha mattes, they require fine trimaps or backgrounds when inferencing. Our CasDGR only needs single RGB images, which is much more convenient for matting applications.

\subsection{Results on Real Image Dataset}

\begin{figure}
\begin{center}
    \includegraphics[width=1\linewidth]{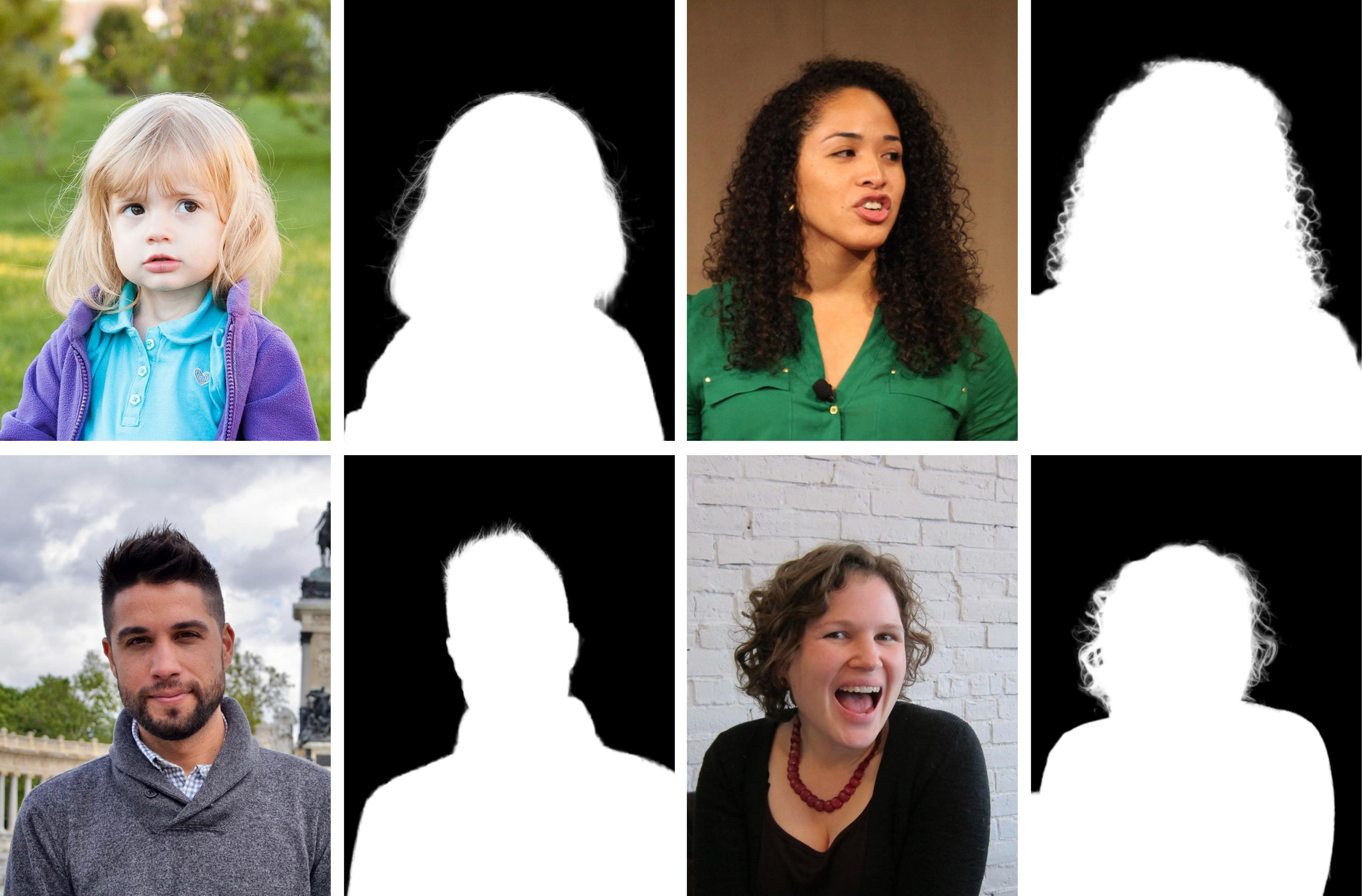}
\end{center}
   \caption{Results on real-world images.}
\label{fig:realworld}
\end{figure}

As a real-world application, the performance on real-world data is significant for image matting methods. To verify the matting effect of our method on real-world images, we construct a real image dataset in the following manner: 1) collection the portrait images from the Internet, and 2) selection some images from a real fashion model dataset constructed by Chen \etal~\cite{SHM2018}. Figure~\ref{fig:realworld} shows 
that our CasDGR model trained on Adobe human image dataset can also produce high-quality alpha mattes on real-world images without the need for additional inputs. 

\section{Conclusions}
In this study, we propose a Cascade Image Matting Network with Deformable Graph Refinement (CasDGR), that can produce high--quality alpha mattes from single RGB images. We adopt the network cascade architecture to progressively refine the foreground details. The proposed DGR module applies GNN on higher--resolution features to further improve matting performance. 
The experimental results on the synthetic dataset and real-world images demonstrate the superiority and generalization of our approach. 


{\small
\bibliographystyle{ieee_fullname}
\bibliography{egbib}
}

\end{document}